\documentclass[letterpaper]{article} 
\usepackage{aaai25}  
\usepackage{times}  
\usepackage{helvet}  
\usepackage{courier}  
\usepackage[hyphens]{url}  
\usepackage{graphicx} 
\usepackage{natbib}  
\usepackage{caption} 
\usepackage{algorithm}
\usepackage{algorithmic}

\usepackage[table]{xcolor}

\usepackage{color}
\usepackage{booktabs} 
 \usepackage{multirow}
\usepackage{colortbl}
\usepackage{cite}
\usepackage{bm}
\usepackage{amsmath,amsfonts}
\usepackage[bottom]{footmisc}
\usepackage[utf8]{inputenc}

\newcommand{\Fmat}[0]{{{\boldsymbol F}}}

\newcommand{\Xmat}{{\boldsymbol X}}

\newcommand{\Zmat}{{\boldsymbol Z}}

\newcommand{\uv}[0]{{\boldsymbol{u}}}

\newcommand{\Xv}{\boldsymbol{X}}

\usepackage{newfloat}
\usepackage{listings}
\DeclareCaptionStyle{ruled}{labelfont=normalfont,labelsep=colon,strut=off} 
\floatstyle{ruled}
\newfloat{listing}{tb}{lst}{}
\floatname{listing}{Listing}
\title{High-Speed FHD Full-Color Video Computer-Generated Holography}
\author{
    Haomiao Zhang\textsuperscript{\rm 1,2}, Miao Cao\textsuperscript{\rm 1,2}, Xuan Yu\textsuperscript{\rm 3}, Hui Luo\textsuperscript{\rm 4},Yanling Piao\textsuperscript{\rm 2}, Mengjie Qin\textsuperscript{\rm 2}, \\Zhangyuan Li\textsuperscript{\rm 1,2}, Ping Wang\textsuperscript{\rm 1,2},  Xin Yuan\textsuperscript{\rm 2}\thanks{Corresponding author.}\\
}
\affiliations{
    \textsuperscript{\rm 1}Zhejiang University.
    \textsuperscript{\rm 2}School of Engineering, Westlake University.\\
    \textsuperscript{\rm 3}Wuhan National Laboratory for Optoelectronics and School of Optical and Electronic Information, Huazhong University of Science and Technology.\\
    \textsuperscript{\rm 4} State Key Laboratory of Optical Field Manipulation Science and Technology, Institute of Optics and Electronics, CAS.\\
    xyuan@westlake.edu.cn
}

\usepackage{bibentry}

\begin{document}

\maketitle

\begin{abstract}
Computer-generated holography (CGH) is a promising technology for next-generation displays. However, generating high-speed, high-quality holographic video requires both high frame rate display and efficient computation, but is constrained by two key limitations: ($i$) Learning-based models often produce over-smoothed phases with narrow angular spectra, causing severe color crosstalk in high frame rate full-color displays such as depth-division multiplexing and thus resulting in a trade-off between frame rate and color fidelity. ($ii$) Existing frame-by-frame optimization methods typically optimize frames independently, neglecting spatial-temporal correlations between consecutive frames and leading to computationally inefficient solutions. To overcome these challenges, in this paper, we propose a novel high-speed full-color video CGH generation scheme. First, we introduce Spectrum-Guided Depth Division Multiplexing (SGDDM), which optimizes phase distributions via frequency modulation, enabling high-fidelity full-color display at high frame rates. Second, we present HoloMamba, a lightweight asymmetric Mamba-Unet architecture that explicitly models spatial-temporal correlations across video sequences to enhance reconstruction quality and computational efficiency. Extensive simulated and real-world experiments demonstrate that SGDDM achieves high-fidelity full-color display without compromise in frame rate, while HoloMamba generates FHD (1080p) full-color holographic video at over 260 FPS, more than 2.6$\times$ faster than the prior state-of-the-art Divide-Conquer-and-Merge Strategy.

\end{abstract}

\section{Introduction}

\begin{figure}[t]
  \centering
   \includegraphics[width=1\linewidth]{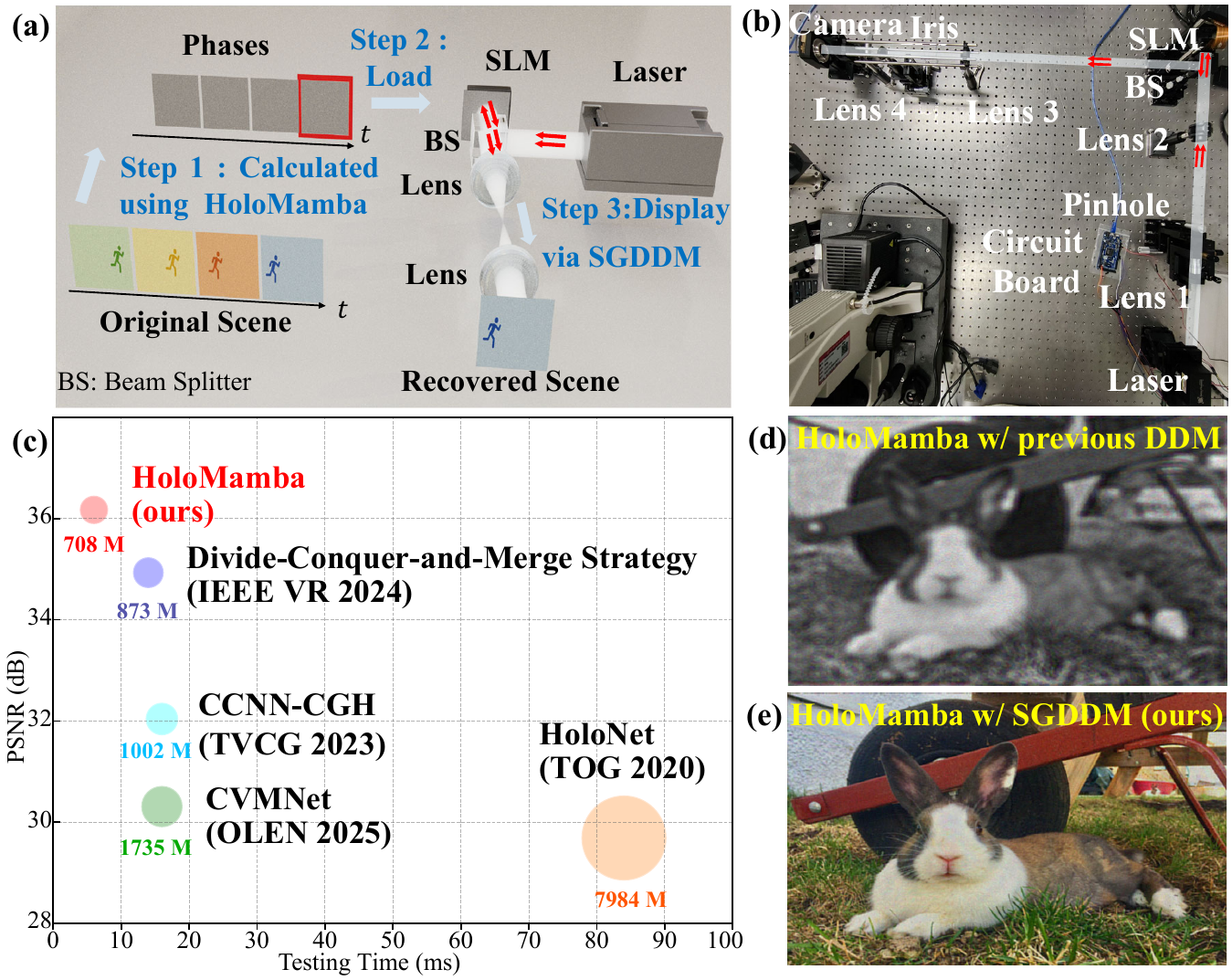}
   \caption{(a) Schematic of CGH model. (b) Real world experimental setup. (c) Comparison of reconstruction quality, testing time and memory requirement with recent deep learning based algorithms. (d,e) Standard DDM fails to perform full-color display with grayscale phase hologram generated by HoloMamba due to the severe color crosstalk. In contrast, our SGDDM preserves high color fidelity.
}
   \label{fig:optical}
\end{figure}

Computer-generated holography (CGH) has emerged as an innovative display technology enabling digital synthesis of both real-world and virtual scenes without physical optical constraints~\cite{brown1966complex}. It holds significant potential in a wide range of applications, including storage~\cite{cheriere2025holographic}, encryption~\cite{fang2020orbital} and display~\cite{xiong2021holographic,gopakumar2024full}. As illustrated in Fig.~\ref{fig:optical}(a), a typical CGH pipeline consists of three key steps. First, holograms consisting of phase values are calculated by a reconstruction algorithm from the target intensity information. Next, the holograms are sequentially loaded onto devices such as spatial light modulators (SLMs) to modulate the incident light fields. Finally, the modulated light fields propagate through free space and reconstruct the target information at the display plane, which can be perceived directly by human observers or captured by a camera. Practical CGH displays demand high-speed full-color reconstruction, which requires both high frame rate display strategies and efficient algorithms.

To achieve {\em high frame rate full-color display}, depth-division multiplexing (DDM) offers a single-shot solution that operates at the native speed of SLM by superposing phase information of the RGB color channels within one hologram~\cite{makowski2008colorful}. However, the effectiveness of DDM depends on the optical characteristics of the hologram. Learning-based algorithms tend to generate over-smoothed phase distributions, resulting in limited angular spectral and degraded depth selectivity~\cite{dong2025motion}. Optically, smooth phase distributions result in an extended depth of field (DOF), which causes reconstructed RGB images to overlap along the axial direction, leading to severe color crosstalk and significant degradation in color fidelity, as shown in Fig.\ref{fig:optical}(d). To eliminate inter-channel crosstalk, time multiplexing (TM) is widely employed, where RGB lasers are sequentially switched, and each is synchronized with its corresponding hologram displayed on the SLM~\cite{yang2025high,li2025real}. However, TM suffers from a threefold reduction in frame rate, limiting its applicability to dynamic scenes. Therefore, challenge lies in developing a solution that can {\em simultaneously achieve \underline{high temporal resolution} and strong color fidelity in \underline{full-color} holographic} displays.

In parallel, {\em efficient CGH algorithms} are crucial for high-speed applications. Traditional iterative methods typically rely on techniques such as alternating projection~\cite{gerchberg1972practical}, stochastic gradient descent~\cite{chen2019gradient}, and non-convex optimization~\cite{candes2015phase,candes2015phase2}. These methods are computationally inefficient due to complex optimization procedures. Although existing deep learning approaches have greatly improved the quality and speed of hologram generation, they focus on static hologram generation~\cite{yu2025use}. Specifically, CNN-based structures are widely adopted for their ability to model local wavefront efficiently~\cite{zhong2023real2,li2025real}. More recently, Transformer~\cite{dong2023vision} and Mamba-based architectures~\cite{yang2025high} have demonstrated strong performance in single-frame CGH tasks. Yet, when extended to {\em video CGH}, these methods treat each frame independently, which limits efficiency and ignores temporal correlations across frames. Thus a critical challenge remains in designing a {\em lightweight network} that can efficiently address spatial-temporal modeling for high-speed, high-quality video CGH reconstruction. In other video-related tasks, 3D CNNs suffer from suboptimal performance due to limited receptive fields~\cite{tran2015learning,chang2019free}, and Transformers have high computational costs~\cite{bertasius2021space,arnab2021vivit}. In contrast, Mamba employs state space models (SSMs) to capture long-range dependencies with linear complexity, demonstrating high efficiency in long-sequence modeling tasks~\cite{li2024videomamba,hu2025exploiting}, showing promise for real-time video hologram generation.

In a nutshell, we focus on two key design considerations: $(i)$ Suppress color crosstalk in learning-based DDM by reducing phase smoothness and promoting angular spectral diversity across color channels. To this end, we design an effective spectrum-guided modulation strategy dubbed SGDDM in the frequency domain to mitigate color crosstalk, the result is shown in Fig.~\ref{fig:optical}(e). $(ii)$ Design of an efficient spatial-temporal modeling network for high-speed video CGH reconstruction. Motivated by the effectiveness of Mamba in video-related tasks, we propose an efficient network dubbed HoloMamba based on an asymmetric U-Net backbone, integrating both kernel convolution for robust local feature extraction and Mamba modules for efficient long-range spatial-temporal dependency modeling. In this paper, we first adapt and extend this local-global modeling strategy into video CGH, optimizing its capability for comprehensive spatial-temporal representation for high-speed, high-quality reconstruction. Furthermore, we integrate a bidirectional scanning strategy for joint forward and backward temporal modeling. Our main contributions are listed as follows:
\begin{itemize}
\itemsep0em
    \item We propose a high-speed, full-color video CGH scheme. Furthermore, we build a {\em real world holographic display system} based on a phase-only SLM to verify our framework's ability to simultaneously achieve high color fidelity via SGDDM and efficient spatial-temporal video modeling through the HoloMamba network.

    \item We design an effective \underline{s}pectrum-\underline{g}uided \underline{d}epth \underline{d}ivision \underline{m}ultiplexing (SGDDM), which optimizes phase distribution through frequency modulation and ensures accurate color control, achieving simultaneous full-color holographic display without sacrificing frame rate.

    \item We propose HoloMamba, an end-to-end lightweight network capable of generating high-speed, high-quality full-color video CGH sequences simultaneously. To our best knowledge, HoloMamba is the first framework to unify efficiency, dynamic spatial-temporal modeling, and high-speed color display in FHD video CGH.

\end{itemize}

\section{Related Work}
\label{sec:work}

\subsection{Full-Color CGH Display}
Achieving compact and efficient full-color holographic displays is vital in practical applications. Spatial multiplexing employs three SLMs to improve reconstruction quality but introduces complex optical paths and higher costs~\cite{piao2019chromatic}. Time multiplexing synchronizes SLM/laser timing for color display using a single SLM, but triples time consumption~\cite{choi2022time}. Frequency multiplexing utilizes one SLM while compromising resolution to simple scenes and requiring additional spatial filtering, thereby increasing optical complexity~\cite{kozacki2016color,lin2019single}. In contrast, depth-division multiplexing (DDM) assigns each color channel (R, G, B) to a distinct focal plane, enabling simultaneous full-color reconstruction on a single SLM~\cite{markley2023simultaneous}. By avoiding sequential display, DDM preserves temporal resolution and simplifies the optical system, making it well-suited for high frame rate full-color holographic applications. Typically, DDM employs iterative optimization to gradually refine and superimpose the phase information for each RGB channel~\cite{kim2023multi}. However, DDM faces a key challenge when combined with deep learning algorithms because networks often produce over-smoothed phase distributions with extended DOF, causing color crosstalk in full-color CGH display.

\subsection{Efficient CGH Algorithms}

Deep learning have greatly improved the quality and efficiency of CGH, marking a significant step toward real-time, high-quality holographic displays~\cite{shui2022diffraction,liu20234k}. Specifically, Peng et al. introduced a novel CGH architecture named HoloNet, which enables real-time 2D holographic displays with approximately 64 ms for FHD hologram generation~\cite{peng2020neural}. Shi et al. proposed a residual network architecture that efficiently synthesizes photorealistic full-color 3D holograms in about 300 ms~\cite{shi2021towards,shi2022end}. Zhong et al. employed lightweight Complex-CNNs (CCNNs) to achieve the same goal, with both approaches reducing GPU memory usage by pruning network parameters, resulting in a generation time of 16 ms for FHD grayscale images~\cite{zhong2023real}. Dong proposed a divide-and-conquer strategy combined with a merging approach to address the challenges of limited memory and computational capacity in CGH generation, with an acceleration of up to 3$\times$ and 2$\times$ compared with HoloNet and CCNNs respectively~\cite{dong2024divide}. With long-range modeling capability and relatively high computational efficiency, Mamba-based CVMNet effectively reduces the number of parameters while generating FHD high-quality holograms in just 16 ms~\cite{yang2025high}. Yet, these algorithms are inefficient for video hologram generation, as they optimize each frame independently without accounting for temporal dependencies, making them impractical for real-world applications.

\section{Preliminaries of CGH model}

As shown in Fig.~\ref{fig:optical}(a), for any given frame of a video sequence in a holographic display system, the corresponding optical field modulated by the phase-only SLM is $\uv_{\textrm{SLM}}=\exp (i\phi(x,y))$, where $\phi(x,y)$ denotes the loaded phase pattern. Then the intensity on the target plane $I_{target}$ is obtained by the angular spectrum method (ASM) as:
\begin{equation}
\small{I_{\textrm{target}}(x,y) = \left|\mathcal{F}^{-1}\{ H(f_x,f_y)  \mathcal{F}\{e^{i\phi(x, y)}\} \} \right|^2},
\label{eqn: 1}
\end{equation}
where $H(f_x,f_y)={\rm exp}[{ikz\sqrt{1-(\lambda f_x)^2-(\lambda f_y)^2}}]$ is the transfer function,  $k=2\pi/\lambda$ represents the wave number, and $\lambda$ is the wavelength; $f_x$ and $f_y$ are the spatial frequencies in the $x$ and $y$ directions respectively. $\mathcal{F}\left \{  \cdot \right\}$ and $\mathcal{F}^{-1}\left \{\cdot \right\}$ denote the Fourier and inverse Fourier transform respectively. This framework forms the mathematical backbone of CGH reconstruction and supports multiple multiplexing strategies (e.g., TM, DDM) for full-color holography.

\begin{figure}[h]
  \centering
   \includegraphics[width=0.9\linewidth]{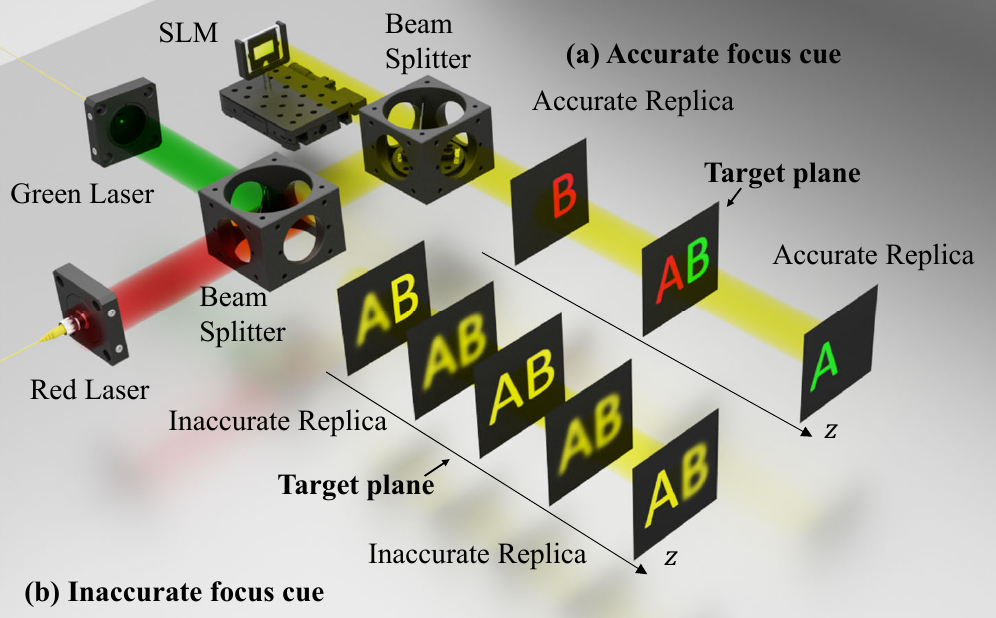}
   \caption{Illustration of depth replica in DDM.
}
   \label{fig:depthrep}
\end{figure}

\noindent{\bf Learning-based CGH in Constrained Optimization:}
CGH is formulated as an ill-posed inverse problem aiming to find a phase distribution $\phi_{\text{opt}}(x,y)$ such that the propagated intensity matches the ground truth image $I_{\text{gt}}$:
\begin{equation}
\textstyle \phi_{\text{opt}}(x,y) = \arg\min_{\phi} \mathcal{L}\left(I_{\textrm{target}}, I_{\text{gt}}\right),
\label{eqn: 2}
\end{equation}
where $\mathcal{L}$ is the loss function. Notably, this formulation relies solely on the fidelity of intensity, without incorporating explicit constraints on the spectral or other physical properties. With such naive supervision, it often yields over-smoothed phase solutions, especially when the phase is initialized with a uniform (low-frequency) or zero phase~\cite{sui2024non}. The relationship between a phase distribution and its angular spectrum can be described by Parseval’s theorem:
\begin{equation}
\textstyle
\small{\iint \left| \nabla\phi(x, y) \right|^2 dx dy = \iint \left( f_x^2 + f_y^2 \right) \left| \Phi(f_x, f_y) \right|^2 df_x df_y,}
\label{eqn:freq2d}
\end{equation}
where $\Phi(f_x,f_y)$ is the Fourier spectrum of $\phi(x,y)$. Due to the limited total gradient energy of a smooth phase distribution, the spectral energy must rapidly decay in the high-frequency region. The relationship between diffraction angle $\theta$ and spatial frequency can be described by $\sin\theta = \lambda \sqrt{f_x^2 + f_y^2}$, indicating that low-frequency components diffract at smaller angles. As a result, the reconstructed intensity remains relatively invariant over an extended axial range with larger DOF, which can be particularly problematic in color holography. 
When applied to DDM, the wavelength-dependent transfer function $H(f_x,f_y)$ leads to depth replicas. As shown in Fig.~\ref{fig:depthrep}, a phase optimized for a green-light target at one depth can produce an in-focus replica at another depth under red illumination. In large-DOF systems, these visually indistinguishable replicas remain sharp and overlap with the target across a wide range, causing unavoidable color–depth ambiguity and degrading color fidelity. By displaying each color channel sequentially, TM avoids inter-channel crosstalk at the cost of temporal resolution, limiting its applicability in high-speed full-color display. More detailed theoretical analysis can be found in Sections 1 and 2 of Supplementary Material (SM).

\begin{figure*}[h]
  \centering
   \includegraphics[width=0.95\linewidth]{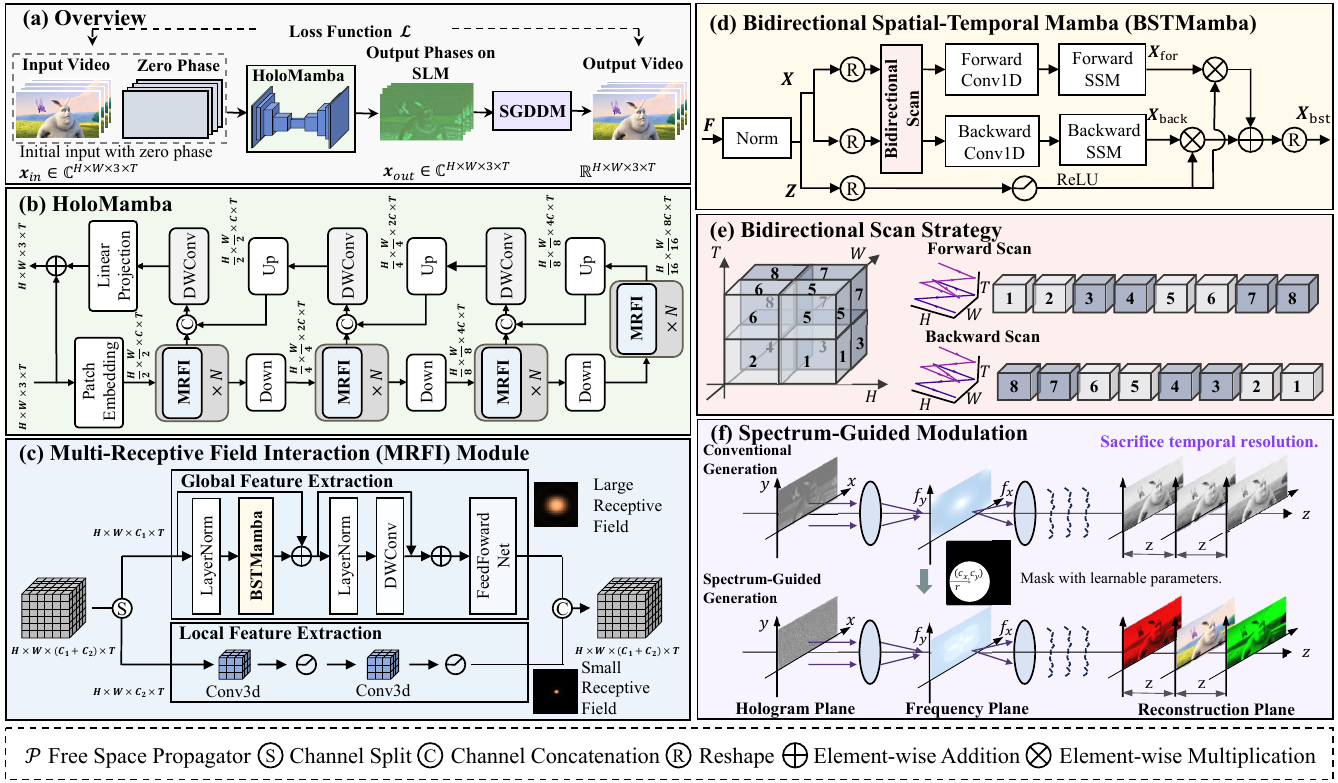}
   \caption{(a) Flowchart of our proposed framework for full-color video CGH reconstruction. (b) The overall network architecture of the proposed HoloMamba. (c) Multi-receptive field interaction (MRFI) module. (d) Bidirectional spatial-temporal mamba (BSTMamba) layer. (e) Bidirectional mamba scanning strategy. (f)  Schematic of spectrum-guided modulation.
}
   \label{fig:structure}
\end{figure*}

\section{Proposed Scheme}
In this section, we present a unified scheme designed for high-speed full-color video CGH generation, composed of SGDDM and HoloMamba, with a particular focus on: ($i$) guiding the network to expand angular spectrum for high-fidelity color reconstruction without sacrificing temporal resolution, ($ii$) overcoming inefficiency and redundancy in per-frame CGH optimization by jointly modeling spatial-temporal correlations across video sequences. 

\subsection{Overall Architecture}

As illustrated in Fig.~\ref{fig:structure}(a), we treat the intensity of RGB frames as the initial amplitude, while the initial phase is uniformly set to zero, forming a complex-valued input $\Xv_{in} \in \mathbb{C}^{H \times W \times 3 \times T}$, where $H$, $W$, and $T$ denote the height, width and frame of the video respectively. The phase-only hologram $\Xv_{out} \in \mathbb{C}^{H \times W \times 3 \times T}$ is estimated by HoloMamba, a lightweight and efficient reconstruction network that ensures spatial fidelity and temporal coherence across full-color video sequences. The output phase is then passed through our SGDDM to generate the final intensity output.

\begin{figure*}[h]
  \centering
   \includegraphics[width=0.85\linewidth]{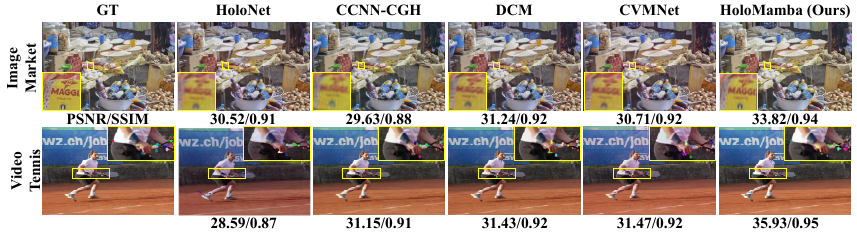}
   \caption{\textit{Numerical Reconstruction Results.} Comparison on full-color FHD image and video datasets. Inter-frame optical flow is visualized for video results to highlight temporal consistency. Zoom in for better view.}
   \label{fig:simulation1}
\end{figure*}

\subsection{HoloMamba Structure}
We design HoloMamba as a three-level asymmetric U-Net for efficient CGH reconstruction, as shown in Fig.\ref{fig:structure}(b). The patch embedding module adopts a cascaded structure with two 3D convolutions (kernel sizes 3$\times$3$\times$3 and 1$\times$1$\times$1), each followed by LeakyReLU\cite{maas2013rectifier}. A spatial stride of 2 reduces resolution by half while maintaining temporal continuity, enabling a hierarchical projection into a high-dimensional latent space through spatial-temporal filtering. At the decoder output, a linear projection comprising two 3D convolutions (1$\times$1$\times$1 and 3$\times$3$\times$3) reconstructs the final output. The encoder has three layers with \underline{m}ulti-\underline{r}eceptive \underline{f}ield \underline{i}nteraction (MRFI) modules and downsampling. MRFI is designed to effectively and efficiently extract features from different scales. The decoder includes three layers with residual depthwise convolution and upsampling. Skip connections are implemented via channel-wise concatenation between corresponding encoder-decoder layers.

\noindent{\bf Multi-Receptive Field Interaction Module:} 
As illustrated in Fig.~\ref{fig:structure}(c), MRFI module adopts a hybrid CNN–Mamba architecture, which partitions the input features into two parallel branches along the channel dimension: ($i$) global receptive feature extraction ($C_1$ channels) leverages Mamba's linear-complexity state-space modeling for long-range dependency capture. ($ii$) Local receptive field CNN ($C_2$ channels) with kernel-level attention to extract spatially localized features. This dual-branch design integrates Mamba's efficient global modeling with CNN's local feature extraction capabilities, achieving a good trade-off between computational efficiency and reconstruction quality. 

The local receptive field CNN branch employs a multi-scale 3D convolutional hierarchy to extract local features, operating in parallel with the global receptive branch. Specifically, it applies 3$\times$3$\times$3 and 1$\times$1$\times$1 kernel to balance detail extraction and computational efficiency.

In contrast, the global receptive branch is responsible for modeling long-range spatial-temporal dependencies. As shown in Fig.~\ref{fig:structure}(c), this module is composed of layer normalization, \underline{b}idirectional \underline{s}patial-\underline{t}emporal \underline{m}amba (BSTMamba), depthwise convolution (DWConv), and feed forward network (FFN). We reshape the video data $\Xv^{(k)}\in \mathbb{C}^{B\times C_1  \times T \times H\times W}$ into one-dimensional sequences $\Xv^{(k)}\in \mathbb{C}^{B\times C_1  \times (T \times H\times W)}$. BSTMamba layer then captures forward-backward contextual relationships using selective SSMs, followed by a DWConv layer (kernel size 3$\times$3$\times$3) to preserve fine-grained details. The operations are formulated as: 
\begin{equation}
\small{
\begin{aligned}
   &\Xv^{(k)}=\textrm{BSTM}({\textrm{LN}}(\Xv^{(k-1)}))+\Xv^{(k-1)}, \\
   &\Xv^{(k)}=\textrm{DWConv}(\textrm{LN}(\Xv^{(k)}))+\Xv^{(k)},
\end{aligned}\label{global}
}
\end{equation}
where $\textrm{LN}$ refers to the layer normalization. The output is then reshaped to $\Xv^{(k)}\in \mathbb{C}^{B\times C_1  \times T \times H\times W}$. FFN aims to improve model's non-linearity and thus enhance its representation ability.

\noindent{\bf BSTMamba Block:} 
BSTMamba block is designed for efficient spatial-temporal information modeling. As shown in Fig.~\ref{fig:structure}(d), the input feature $\Fmat^{(k-1)}$  first passes through two separate linear layers followed by SiLU activations that splits the data into two parallel paths $\Xmat^{(k)}$ and $\Zmat^{(k)}$. $\Xmat^{(k)}$ is processed by a 1D convolution followed by a $\mathrm{ForwardSSM}$ and layer normalization to produce the forward context $\Xmat_{\mathrm{For}}^{(k)}$, while a similar pipeline $\mathrm{BackwardSSM}$ computes the backward context $\Xmat_{\mathrm{Back}}^{(k)}$ using the same $\Xmat^{(k)}$. These two directional representations are modulated by $\Zmat^{(k)}$ through element-wise multiplication and summed. The final fused output $\Xmat_{\mathrm{bst}}^{(k)}$ is obtained via a linear projection. The process can be formulated as follows, where $\mathrm{Lin}$ represents the linear layer and $\odot$ denotes the Hadamard product:  
\begin{equation}
\small{
\begin{split}
       \Xmat^{(k)} &= \mathrm{SiLU}(\mathrm{Lin}(\Fmat^{(k-1)})),\\ \Zmat^{(k)} &= \mathrm{SiLU}(\mathrm{Lin}(\Fmat^{(k-1)})), \\
   \Xmat_{\mathrm{For}}^{(k)} &= \mathrm{LN}(\mathrm{ForwardSSM}(\mathrm{Conv1d}(\Xmat^{(k)}))), \\
   \Xmat_{\mathrm{Back}}^{(k)} &= \mathrm{LN}(\mathrm{BackwardSSM}(\mathrm{Conv1d}(\Xmat^{(k)}))), \\
   \Xmat_\mathrm{bst}^{(k)} &= \mathrm{Lin}(\Xmat_{\mathrm{For}}^{(k)} \odot \Zmat^{(k)} + \Xmat_{\mathrm{Back}}^{(k)} \odot \Zmat^{(k)}).
\end{split}
}
\end{equation}\label{bstm}

\noindent{\bf Bidirectional Scan Strategy:} As shown in Fig.~\ref{fig:structure}(e), $\mathrm{ForwardSSM}$ handles the input 1D sequence along the spatial-temporal axis in the order of Height, Width and Time, while $\mathrm{BackwardSSM}$ applies the same operation on the reversed sequence. This design maintains the spatial continuity within each frame while enabling the model to access information from both past and future frames.  

\subsection{Spectrum-Guided Depth-Division Method}

In SGDDM, we introduce an explicit spectrum-guided modulation mask $\mathcal{M}_{C}(f_x, f_y), C \in \{R, G, B\}$ into the Fourier domain as:
\begin{equation}
\scalebox{0.88}{
$I_{\textrm{target}}(x,y) = \left|\mathcal{F}^{-1}\{ H(f_x,f_y) \mathcal{M}_C(f_x, f_y) \mathcal{F}\{e^{i\phi(x, y)}\} \} \right|^2$,}
\label{eq:sgddm_prop}
\end{equation}
where we recall that  the optical field modulated by the phase-only SLM is $\uv_{\rm SLM} =\exp(i\phi(x, y))$, with $\phi(x, y)$ being the loaded phase pattern.

As shown in Fig.~\ref{fig:structure}(f), each $\mathcal{M}_C$ is parameterized as a circular binary filter centered at $(c_x, c_y)$ with radius $r$, where $c_x$, $c_y$, and $r$ are learnable parameters. Physically, applying such an off-axis mask in the frequency domain is equivalent to introducing a linear phase ramp in the spatial domain:
\begin{equation}
\small{
\phi_{\text{eq}}(x, y) = \phi_{\text{ori}}(x, y) + 2\pi(c_x x + c_y y),}
\label{eq:linear_phase}
\end{equation}
where $\phi_{\textrm{ori}}(x,y)$ denotes the original network output. This modulation shifts the angular spectrum and increases the phase gradient energy:
\begin{equation}
\small{
\nabla \phi_{\text{eq}}(x, y) = \nabla \phi_{\text{ori}}(x, y) + 2\pi(c_x, c_y).}
\label{eq:grad_boost}
\end{equation}

According to Parseval’s theorem in Eq.~\eqref{eqn:freq2d}, this spectral shift effectively broadens the angular bandwidth, thereby increasing the numerical aperture (NA) and reducing the depth of field (DOF). For the $i$-th wavelength, the DOF can be approximated as $\mathrm{DOF}_i \approx {\lambda_i}/{\mathrm{NA}_i^2}$. To minimize inter-channel crosstalk, the axial distance between adjacent focal planes should satisfy the following requirement:
\begin{equation}
\textstyle
|z_i - z_j| > \frac{1}{2} \left( \frac{\lambda_i}{\mathrm{NA}_i^2} + \frac{\lambda_j}{\mathrm{NA}_j^2} \right), \quad i \ne j.
\label{eq:depth_sep}
\end{equation}
This condition guides the learning of $\mathcal{M}_C$ to ensure depth-wise separation of RGB channels.

To enable end-to-end training with non-differentiable binary masks, we adopt a continuous surrogate strategy. Specifically, we approximate each binary mask with a soft circular function during training:
\begin{equation}
\small{
\widetilde{\mathcal{M}}_C(f_x, f_y) = \sigma\left(\tau \cdot \left[ r^2 - (f_x - c_x)^2 - (f_y - c_y)^2 \right] \right),}
\label{mask}
\end{equation}
where $\sigma(\cdot)$ is the sigmoid function and $\tau$ controls the sharpness of the transition. During the forward pass, the hard binary mask is used to simulate physical masking, while gradients are back-propagated through the soft $\widetilde{\mathcal{M}}_C$ to update the parameters $(c_x, c_y, r)$. We initialize $\tau$ at 0.00625 and double it after each training epoch. To ensure numerical stability, we set $\tau$ at a maximum value of 1.6 in our experiments. This ensures soft mask to gradually approximate an ideal binary boundary, bridging physical realism and differentiability.

\subsection{Loss Function}
 
Hereby, we introduce a hybrid loss function designed for holographic reconstruction.

\noindent{\bf MSE loss:} Mean square error (MSE) between the desired video frames $ \left\{\mathbf{y}_t \right\}^T _{t=1} \in \mathbb{R} ^{n_x \times n_y}$ and the reconstructed complex amplitude $\hat{\mathbf{x}}_t$ can be written as:
\begin{equation}
\small{
    {\mathcal{L}_{\rm MSE}} = \textstyle \dfrac{1}{Tn_xn_y}\sum_{t=1}^T \Vert \mathbf{y}_t-|\mathcal{P}(\hat{\mathbf{x}}_t)| \Vert^2_2,
    }
\end{equation}
where $n_xn_y$ is the total number of pixels, $\mathcal{P\{\cdot\}}$ denotes the free-space propagation in Eq.~\eqref{eq:sgddm_prop}.

\noindent{\bf FFL:} Focal frequency loss (FFL) emphasizes the recovery of crucial high-frequency components that are often difficult to reconstruct, which is defined as:
\begin{equation}
\small{
{\mathcal{L}_{\rm FFL}} = \textstyle \dfrac{1}{Tn_xn_y}\sum_{t=1}^T w \cdot \Vert \mathcal{F}(\mathbf{y}_t)-\mathcal{F}(|\mathcal{P}(\hat{\mathbf{x}}_t)| )\Vert^2_2,
}
\end{equation}
where $w = \left|\mathcal{F}(\mathbf{y}_t) - \mathcal{F}(|\mathcal{P}(\hat{\mathbf{x}}_t)|)\right|^{\alpha}$ is a soft frequency-aware regularization weight, implemented as a dynamic weighting matrix in the Fourier domain. This formulation amplifies the penalty for reconstruction errors in high-frequency regions, and the superscript $\alpha$ controls the sensitivity to frequency-domain discrepancies.

Finally, the overall loss function $\mathcal{L}$ can be written as:
\begin{equation}
\small{
\mathcal{L}=\lambda_{\rm MSE} \cdot \mathcal{L}_{\rm MSE}+ \lambda_{\rm FFL} \cdot \mathcal{L}_{\rm FFL},
\label{eqn: loss}
}
\end{equation} 
where $\lambda_{\rm MSE}$,  $\lambda_{\rm FFL}$ are constants to balance the two terms. Detailed loss settings are provided in Section 5 of the SM.

\section{Experiments}
\label{sec:experiment}

In this section, we evaluate the quality of HoloMamba and SGDDM through simulations and real-world experiments. 

\subsection{Implementation Details}
We use Pytorch 2.0 and Python 3.9 trained for 50 epochs with an NVIDIA RTX 8000 GPU. We adopt Adam as the optimizer~\cite{kingma2014adam} with a learning rate of $1e-4$. The {\tt DAVIS2017} dataset~\cite{pont20172017} is used to evaluate the trained models. The Peak Signal-to-Noise Ratio (PSNR) and Structured Similarity Index Metric (SSIM)~\cite{wang2004image} are employed to assess the reconstructed results. We perform full-color image experiments on the validation set of {\tt DIV2K}~\cite{Lim_2017_CVPR_Workshops} and full-color video experiments on the test set of {\tt DAVIS2017}.


\begin{table}[h!]
\centering
\setlength{\tabcolsep}{1pt}
\begin{tabular}{c|ccccc}
    \hline
    Method  & Param$_\text{(K)}$ & Mem$_\text{(M)}$ & FPS & PSNR/ SSIM
 &Warp\\
    \hline
     HoloNet 
& 2868.7& 7,984& 16&29.69/0.90 &0.054
\\
 CCNN-CGH 
& 42.2& 1,002& 61& 32.01/0.92 &0.032
\\
 DCM& 112.8& 873& 99& 32.83/0.93 &0.048
\\
     CVMNet& 146.9& 1,735& 60.1&30.28/0.90 &0.031
\\

\textbf{HoloMamba}& \textbf{44.7}& \textbf{708}& \textbf{267$\uparrow$}&\textbf{35.44$\uparrow$/0.95$\uparrow$}&\textbf{0.022$\downarrow$}\\
     \hline
\end{tabular}
\caption{Quantitative comparison of recent CGH algorithms on benchmark datasets with FHD resolution. }
\label{tab:PSNR}
\end{table}

\subsection{Simulation Results}

We first compare HoloMamba with HoloNet~\cite{peng2020neural}, CCNN-CGH~\cite{zhong2023real}, Divide-Conquer-and-Merge Strategy~\cite{dong2024divide} (denoted as DCM, using CCNN $\times 4$ configuration), and CVMNet~\cite{yang2025high} using TM for fair comparison. We evaluate both simulated image and video reconstructions. As shown in Fig.\ref{fig:optical}(c) and Tab.\ref{tab:PSNR}, HoloMamba achieves an average PSNR of 35.44 dB and SSIM of 0.95, outperforming all baselines. In terms of efficiency, our method reduces GPU memory consumption by 18.6$\%$ and achieves a 2.6$\times$ speedup over DCM. Furthermore, it exhibits lower warping errors compared to other methods in video scenes. Fig.~\ref{fig:simulation1} presents selected simulation results on both images and video scenes, where our method demonstrates superior visual quality. The visualization of Lucas-Kanade optical flow shows smoother inter-frame motion trajectories and better structural coherence over time. More simulation results on full-color FHD video sequences are provided in Section 4 of the SM.

\begin{figure}[h]
  \centering
   \includegraphics[width=.9\linewidth]{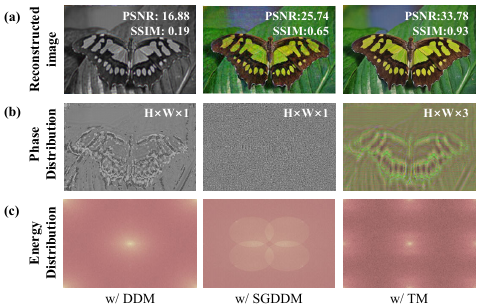}
   \caption{(a) Numerical reconstruction results of full-color images.(b) Phase distribution (c) Light energy distribution of HoloMamba w/ DDM, SGDDM and TM respectively.
}
   \label{fig:pupilsim}
\end{figure}

We further present full-color holographic results using SGDDM in Fig.~\ref{fig:pupilsim}. Without spectral guidance, HoloMamba produces over-smoothed phases with concentrated low-frequency energy, causing severe color crosstalk. In contrast, SGDDM broadens the spectral distribution, enabling accurate color rendering and high-fidelity reconstruction. Additional results on SGDDM’s generalization across baseline networks are provided in Section 4 of the SM.

\begin{figure*}[h]
  \centering
   \includegraphics[width=0.85\linewidth]{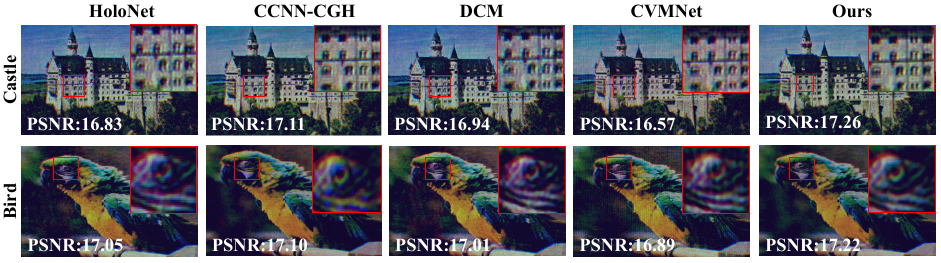}
   \caption{\textit{Real-world data results.} Optical reconstruction results of various algorithms on FHD full-color images based on TM. 
   }
   \label{fig:experiment1}
\end{figure*}

\begin{figure*}[h]
  \centering
   \includegraphics[width=0.85\linewidth]{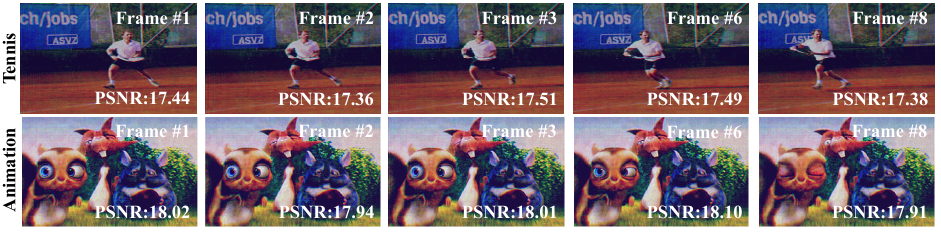}
   \caption{\textit{Real-world data results.} Selected reconstruction frames of our HoloMamba on FHD full-color video frames. 
}
   \label{fig:experiment2}
\end{figure*}

\subsection{Real-World Data Results}
Our real data setup is shown in Fig.~\ref{fig:optical}(b), with extra details given in Section 4 of the SM. To fully validate our method's performance, we conduct three key experiments that demonstrate its capabilities across static reconstruction, dynamic video processing, and full-color display applications.

Our first experiment evaluates and compares the performance of HoloMamba against other deep-learning-based real-time algorithms utilizing TM. As shown in Fig.~\ref{fig:experiment1}, our method produces better reconstruction results with more details. These results demonstrate close alignment with ground truth values, confirming that our network achieves highly effective reconstruction performance while maintaining computational efficiency.

Our second experiment demonstrates HoloMamba’s capability in large-scale full-color video reconstruction. As shown in Fig.~\ref{fig:experiment2}, HoloMamba delivers high-quality reconstructions, with the resulting videos exhibiting both high fidelity and smooth temporal transitions across consecutive frames. Additional results are provided in Section 4 of SM.

\begin{figure}[h]
  \centering
   \includegraphics[width=0.85\linewidth]{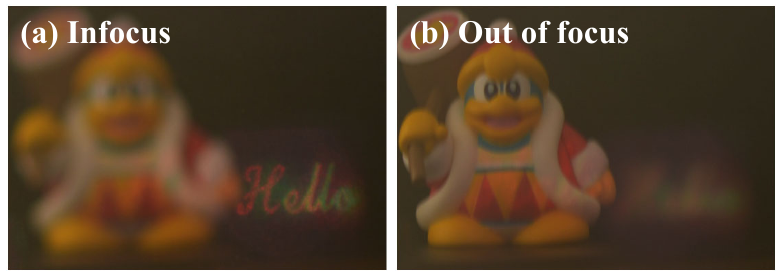}
   \caption{\textit{Real-world data results.} Optical AR experiments. 
}
   \label{fig:experiment3}
\end{figure}

Our third experiment demonstrates full-color holographic results using the proposed SGDDM. As shown in Fig.~\ref{fig:experiment3}, we successfully achieve full-color augmented reality (AR) display from a single grayscale hologram. A toy is positioned at a distance of 45 cm, while the holographic scene is displayed at a distance of 35 cm.

\subsection{Ablation Study}

We conduct ablation studies on the {\tt DAVIS2017} dataset (256$\times$256 resolution, 8 frames) to validate the effectiveness of HoloMamba. More studies on scanning strategies, architecture and loss design are provided in Section 5 of the SM.

\noindent{\bf Comparison with CNN and Transformer:} 
We replace HoloMamba with a 3D CNN~\cite{tran2015learning} and a vision transformer (ViT)~\cite{dosovitskiy2020image} with global self-attention. From Tab.~\ref{tab:CNNtransformer}, we can conclude that HoloMamba achieves high performance with computational efficiency. More detailed analysis of the computational complexity can be found in Section 3 of SM.

\begin{table}[h!]
\centering
  \setlength{\tabcolsep}{4pt}
\begin{tabular}{c|ccc}
    \hline
   Method& Mem$_\text{(M)}$ & FPS& PSNR\\
    \hline
     3D CNN& 13,976& 17& 24.78\\
     ViT& 202,348& 0.20& 29.01\\
HoloMamba& 2,734& 267& 28.74\\
     \hline
\end{tabular}
\caption{Ablation study on the HoloMamba model.}
\label{tab:CNNtransformer}
\end{table}

\noindent{\bf MRFI block:} We analyze the effect of varying the ratio between global (GRFE) and local (LRFE) receptive field extraction blocks. As shown in Tab.~\ref{tab:MRFI}, increasing LRFE reduces both parameters and reconstruction quality. Standalone LRFE (0:1) fails to capture long-range spatial-temporal correlations, while standalone GRFE (1:0) improves quality but increases memory usage. Thus, we adopt a 0.8:0.2 ratio to balance efficiency and quality.

\begin{table}[h!]
\centering
\setlength{\tabcolsep}{3pt}
\begin{tabular}{c|ccccc}
  \hline
  $C_1:C_2$& 0:1&  0.25:0.75& 0.5:0.5& 0.8:0.2&1:0\\
  \hline
 Mem$_\text{(M)}$& 2,028 & 2,274 & 2,499& \textbf{2,734}&2,964\\
 PSNR & 19.44 & 20.41& 23.28& \textbf{28.74}&29.03\\
     \hline
\end{tabular}
\caption{Ablation study on the MRFI block.}
\label{tab:MRFI}
\end{table}

\section{Conclusion}
\label{sec:Conclusion}

We propose a high-speed full-color FHD video CGH scheme to tackle two key challenges: over-smoothed phase that leads to color crosstalk in high frame rate displays, and the lack of spatial-temporal modeling in frame-wise methods. Our proposed SGDDM mitigates color crosstalk by optimizing the angular spectral distribution, enabling one-shot full-color reconstruction without sacrificing frame rate. To further enhance efficiency, we develop HoloMamba, a lightweight asymmetric Mamba-Unet architecture that models spatial-temporal correlations for high-speed phase generation. Extensive simulations and real-world experiments validate the superior performance of our approach, paving a practical way for high frame rate full-color holographic display.

\bibliography{aaai25}

\end{document}